\documentclass[conference]{IEEEtran}
\usepackage{cite}
\usepackage{amsmath,amssymb,amsfonts}
\usepackage{algorithmic}
\usepackage{graphicx}
\usepackage{textcomp}
\usepackage{xcolor}

\usepackage{times}
\usepackage{epsfig}
\usepackage{amsmath}
\usepackage[linesnumbered,ruled,commentsnumbered]{algorithm2e}

\usepackage{bm}
\usepackage{subfigure}
\usepackage{multirow}
\usepackage{hyperref}

\usepackage{booktabs}  

\begin{document}

\title{Progressive Visual Reasoning for Video Captioning Using Weak Annotation}

\author{Jingyi Hou\textsuperscript{1}, Yunde Jia\textsuperscript{1}, Xinxiao Wu\textsuperscript{1}, Yayun Qi\textsuperscript{1}, Yao Hu\textsuperscript{2}\\
\textsuperscript{1}Beijing Laboratory of Intelligent Information Technology, School of Computer Science,\\ Beijing Institute of Technology, Beijing 100081, China\\
\textsuperscript{2}Alibaba Youku Cognitive and Intelligent Lab, Alibaba Group
}

\maketitle

\begin{abstract}
Most existing methods of video captioning are based on strong annotation that refers to annotating videos with paired video-sentences, and annotating such data is both time-consuming and laborious.  It is the fact that there now exist an amazing number of videos with weak annotation that only contains semantic concepts such as actions and objects.  In this paper, we investigate using weak annotation instead of strong annotation  to train a video captioning model. To this end, we propose a progressive visual reasoning method that progressively generates fine sentences from weak annotations by inferring more semantic concepts and their dependency relationships for video captioning. To model concept relationships, we use dependency trees that are spanned by exploiting external knowledge from large sentence corpora. Through traversing the dependency trees, the sentences are generated to train the captioning model. Accordingly, we develop an iterative refinement algorithm to alternate between refining sentences via spanning dependency trees and fine-tuning the captioning model using the refined sentences. Experimental results on several datasets demonstrate that our method using weak annotation is very competitive to the state-of-the-art methods using strong annotation.
\end{abstract}

\section{Introduction}
Deep learning-based video captioning \cite{DBLP:journals/corr/abs-1711-11135,DBLP:journals/tmm/GaoGZXS17,DBLP:conf/cvpr/PanYLM17,DBLP:conf/cvpr/BaraldiGC17,Chen_2018_ECCV,DBLP:journals/corr/abs-1902-10322,DBLP:journals/corr/abs-1906-04375} has achieved impressive progress benefiting from strong annotation datasets with plenty of video and sentence pairs. Yet so far, acquiring sufficient video and sentence pairs  is still a labor-intensive and time-consuming process. It is the fact there are many video datasets with weak annotation that only contains semantic concepts (i.e. actions and objects),  such as object classification datasets and action recognition datasets, and we easily have access to these data for video captioning. Also, it is much easier and less costly to annotate videos with semantic concepts than complete sentences. In this paper, we investigate using weak annotation datasets instead of strong annotation datasets to train a video captioning model.

\begin{figure}[t]
\centering
\includegraphics[width=0.86\columnwidth]{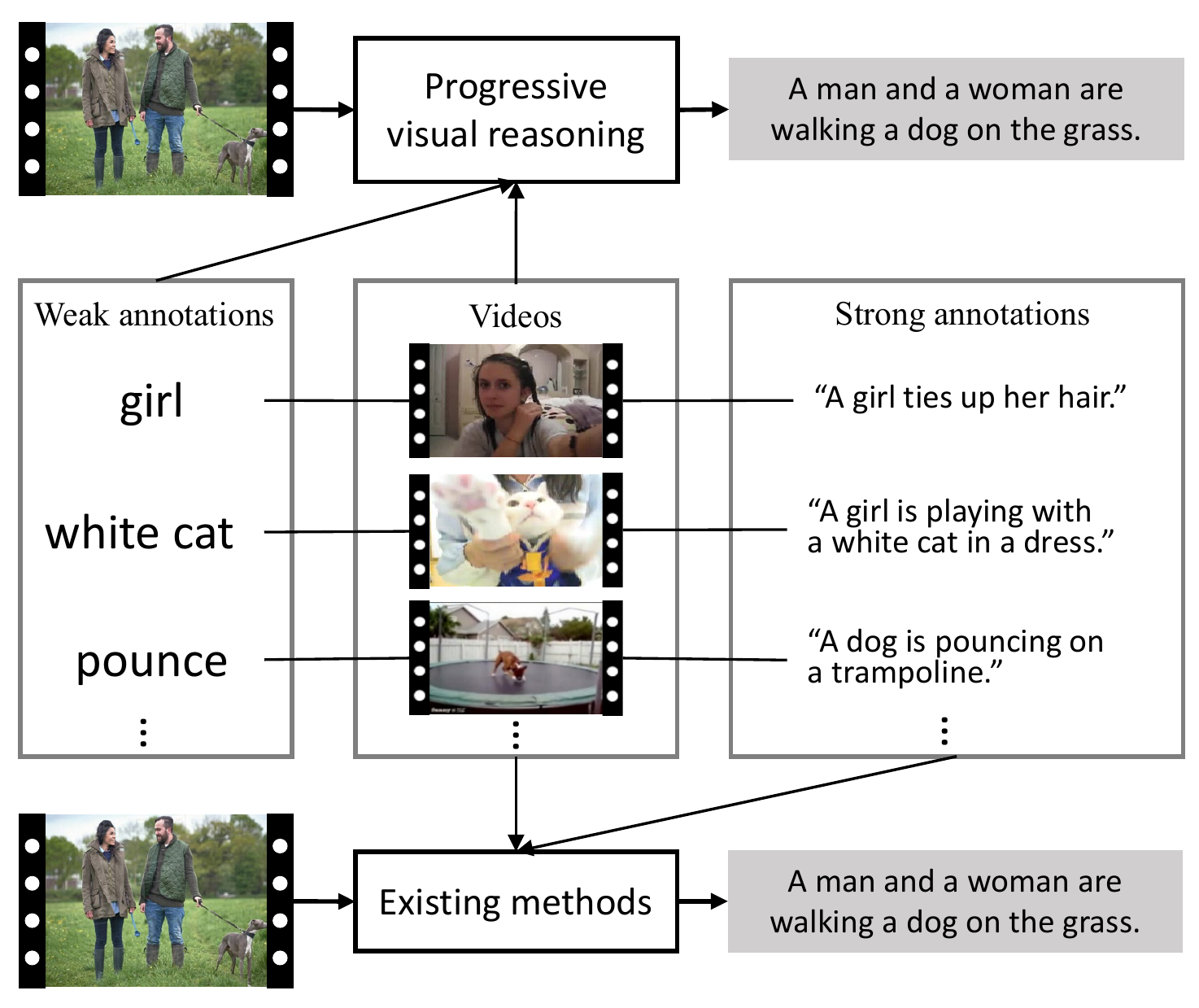}
\caption{Illustration of video captioning tasks using strong and weak annotations.}
\label{traindata}
\end{figure}

Figure~\ref{traindata} shows some examples of using weak annotations to train a video captioning model, where a video recording that “A girl ties up her hair” is annotated with a concept of “girl”. Training a video captioning model using weak annotations faces two problems: (1) how to infer more semantic concepts and their relationships from a given semantic concept, and (2) how to generate sentences\footnote{The generated sentences in training are called \emph{pseudo sentences} that are
distinguished from sentences output by a video captioning model.} for training videos from the inferred semantic concepts and their relationships.

To handle these problems, we propose a progressive visual reasoning method that progressively refines pseudo sentences by inferring more semantic concepts and their dependency relationships.
Specifically, the pseudo sentences are generated by exploiting external knowledge from large sentence corpora and these pseudo sentences are used as ground truth for training a video captioning model.
Inspired by the curriculum learning framework \cite{DBLP:conf/icml/BengioLCW09,DBLP:conf/aaai/MitchellCHTBCMG15} that gradually
increases the difficulty of training samples in an iterative learning
procedure, we iteratively train the captioning model, mimicking the learning strategy of human beings.
For each iteration, the captioning model infers semantic concepts and their dependency relationships to generate pseudo sentences, and then the captioning model is fine-tuned to produce new semantic concepts using the generated pseudo sentences. 
Such an iterative learning enables the captioning model to start with weak annotations and progressively learn to generate and refine pseudo sentences, guiding the captioning model to converge faster and achieve better results \cite{DBLP:conf/icml/BengioLCW09,DBLP:conf/aaai/MitchellCHTBCMG15}.

To model the dependency relationships between semantic concepts of a video, we build dependency trees whose nodes represent semantic concepts and whose edges represent the relationships.
We develop an iterative refinement algorithm that alternates between generating pseudo sentences via spanning dependency trees and fine-tuning the video captioning model using the generated pseudo sentences.
During each iteration, the captioning model generates semantic concepts of videos based on visual grounding, and then dependency trees are spanned to model the dependency relationships between semantic concepts via link prediction learned from external knowledge. The external knowledge is represented by triplets of semantic concepts (concept-relationship-concept) extracted from existing sentence corpora.
The semantic concepts are described by noun phrases and verb phrases, so combining a noun phrase and a verb phrase makes it possible to construct a subject-predicate phrase serving as the root node of the dependency tree.
In order to decide whether to combine a noun phrase and a verb phrase or not, we introduce the spatial consistency of a subject and a predicate as a measurement based on the observation that the predicate is the action taken by the corresponding subject and their regions in a video are spatially aligned.
Since the dependency tree contains all the contextual and semantic information of a sentence, a pseudo sentence can be directly generated from the dependency tree according to the language syntax.
Finally, the generated pseudo sentences associated with the corresponding videos are used to fine-tune the captioning model by propagating the information through the dependency trees and translating them into textual descriptions.
Thus the captioning model is updated and outputs new semantic concepts for the next iteration.

The main contributions of this paper are summarized as
\begin{itemize}
\item We propose a progressive visual reasoning method that progressively refines pseudo sentences by exploiting the external knowledge to train the video captioning model.
\item We make the first attempt to investigate video captioning using weak annotation, and experiments on several datasets demonstrate that our method using weak annotation is very competitive to the state-of-the-art methods with strong annotation. 

\end{itemize}

\section{Related Work}

Early methods of video captioning are mainly based on templates \cite{DBLP:conf/iccv/GuadarramaKMVMDS13,DBLP:conf/aaai/KrishnamoorthyMMSG13,DBLP:conf/coling/ThomasonVGSM14}.
These methods pre-define fixed-structured sentence templates and fill the recognized visual concepts in the templates to generate captions.
Guadarrama et al. \cite{DBLP:conf/iccv/GuadarramaKMVMDS13} develop a language driven method that learns semantic relationships to build semantic hierarchies and uses a web-scale language model to fill in novel verbs.
Krishnamoorthy et al. \cite{DBLP:conf/aaai/KrishnamoorthyMMSG13} first apply visual recognition algorithms to detect objects and actions from videos, and then generate sentences with the structure of subject-verb-object according to the likelihood estimated in real-world and web corpora.
Thomason et al. \cite{DBLP:conf/coling/ThomasonVGSM14} further add scene recognition to enrich the generated descriptions with template subject-verb-object-place by designing a factor graph model.
These template-based methods utilize concept detectors and probabilistic knowledge of extra sentence corpora to generate sentences based on the pre-defined syntactic structures. 
In contrast, our method generates arbitrary sentences with diverse structures via  progressive visual reasoning and guarantees more accurate captions in two aspects:
(1) the iterative learning strategy to improve the concept recognition performance
and (2) the spatial consistency mechanism to reason dependency relationships between concepts not only according to language prior but also from visual information.

There are increasing studies \cite{DBLP:conf/cvpr/DonahueHGRVDS15,DBLP:conf/naacl/VenugopalanXDRM15,DBLP:conf/iccv/VenugopalanRDMD15,DBLP:conf/cvpr/PanXYWZ16,Chen_2018_ECCV} that integrate video encoding networks and sequence-based decoding networks for video captioning.
Beyond encoding the entire video into a global representation, many methods exploit local visual features or semantic attributes to generate textual descriptions of videos \cite{DBLP:conf/cvpr/YuWHYX16,DBLP:journals/corr/abs-1711-11135,DBLP:journals/tmm/GaoGZXS17,DBLP:conf/cvpr/PanYLM17,DBLP:conf/cvpr/BaraldiGC17,DBLP:conf/mm/LiuRY18,DBLP:journals/corr/abs-1902-10322}.
Yu et al. \cite{DBLP:conf/cvpr/YuWHYX16} use a hierarchical-RNN framework that applies spatial and temporal attention to focus on salient visual elements for video paragraph captioning.
Wang et al.\cite{DBLP:journals/corr/abs-1711-11135} bring hierarchical reinforcement learning with the attention mechanism to capture local temporal dynamics into the video captioning task.
Both focusing on adapting LSTM frameworks, Gao et al. \cite{DBLP:journals/tmm/GaoGZXS17} propose a unified framework contains an attention-based LSTM model with semantic consistency, and Pan et al. \cite{DBLP:conf/cvpr/PanYLM17} present an LSTM that explores semantic attributes in video with a transfer unit.
For video encoding, Baraldi et al. \cite{DBLP:conf/cvpr/BaraldiGC17} introduce a time boundary-aware video encoding scheme that adapts its temporal structure to the input, and Liu et al. \cite{DBLP:conf/mm/LiuRY18} construct a two-branch network to collaboratively encode videos.
Aafaq et al. \cite{DBLP:journals/corr/abs-1902-10322} develop a visual encoding method that utilizes both spatio-temporal dynamics and high-level semantics from the video.
Owing to the success of exploiting semantic relationships to interpret videos, integrating relation reasoning into video captioning models has attracted more and more attentions recently.
Zhang and Peng \cite{DBLP:journals/corr/abs-1906-04375} capture the temporal dynamics on salient object instances and learn spatio-temporal graphs to represent video contents for captioning.
Zhang et al.~\cite{zhangcvpr2020caption} infer the relationships among objects in videos and exploit external language model to alleviate the long-tail problem in video captioning.
Although those aforementioned methods have achieved promising results, they require strong annotation datasets with large amount of paired video-sentence data to train video captioning models. Rather than heavily relying on the complete sentences of videos, our method uses semantic concepts as weak supervision to train the captioning model.

\begin{figure*}[t]
\centering
\includegraphics[width=0.86\textwidth]{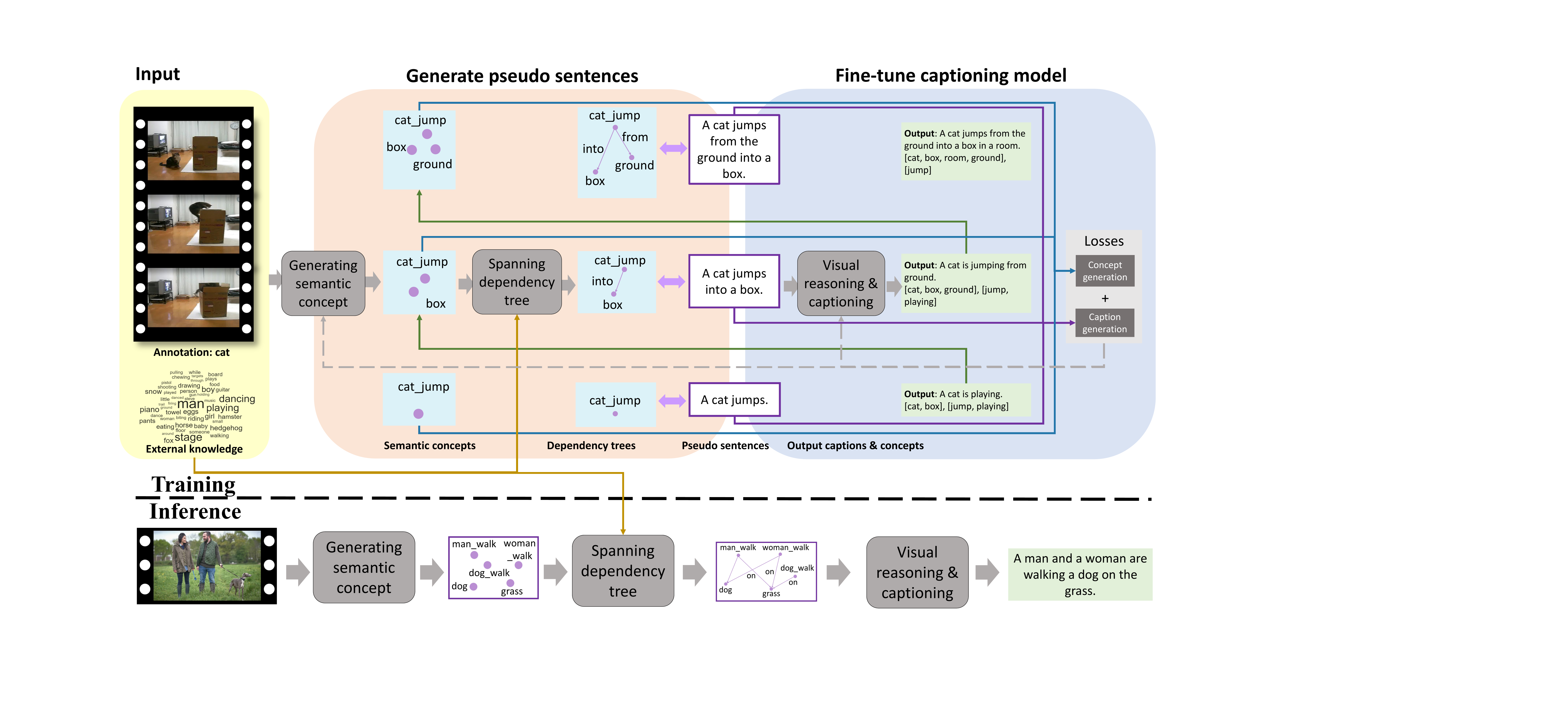}
\caption{Overview of the progressive visual reasoning method. In training phrase, a captioning model is iteratively updated to generate not only the captions but also the semantic concepts by exploiting external knowledge extracted from the large-scale external sentence corpus. The newly generated concepts are fed into the next iteration to make the dependency tree grow. }
\label{task}
\end{figure*}

In recent years, several methods have been proposed for image captioning in unsupervised ways.
Laina et al. \cite{DBLP:journals/corr/abs-1908-09317} use the idea of domain alignment to achieve the goal of unsupervised image captioning by mapping images into a shared embedding space with sentences.
Feng et al. \cite{DBLP:journals/corr/abs-1811-10787} propose an unsupervised image captioning method that uses generative adversarial networks for image and sentence reconstructions with the guidance of semantic concepts derived from a pre-trained detector.
Instead of directly reconstructing images to learn their descriptions, we explicitly exploit the sentence constructions and  propose the progressive visual reasoning to gradually generate pseudo sentences to train the video captioning model.
Moreover, our method leverages easily available semantic concepts from existing object classification and action recognition datasets, without requiring any pre-trained detectors.

\section{Progressive Visual Reasoning}

Our goal is to train a video captioning model using weak annotations, where each training video is annotated by a semantic concept.
To achieve this goal, we propose a progressive visual reasoning method that progressively refines pseudo sentences given a semantic concept annotation to fine-tune the captioning model, as shown in Figure~\ref{task}.
During training, an iterative refinement algorithm is developed to train the video captioning model via the pseudo sentences gradually generated by exploiting the external knowledge.
During testing, given an input video, the caption is generated by the well trained captioning model.
In the following sections, we first introduce our model, and then  describe the training policy by presenting loss function and the detailed iterative refinement algorithm.

\subsection{Our Model}
The overall architecture of our model can be divided into the following steps:
(1) generating semantic concepts from the input video;
(2) spanning the dependency tress based on the semantic concepts and the external knowledge;
(3) visual reasoning and captioning based on the dependency trees.

\subsubsection{Generating Semantic Concept}

The proposed captioning model generates semantic concepts by retrieving all the semantic concepts contained in external sentence corpora and deciding which should be preserved.
The generated semantic concepts are used as the candidate nodes of the dependency trees to be spanned.
For simplicity, only the concepts of objects and actions are considered, and the concepts of objects and actions  correspond to noun phrases and verb phrases in the captions, respectively.
We define an indicator vector $\bm{g}^{\rm{k}} \in
 \mathbb{R}^{{N}^{\rm{k}} \times 1}$
where $\rm{k} \in \{\rm{o},\rm{a}\}$ and ${N}^{\rm{k}}$ represents the total number of objects or actions in external sentence corpora. ``$\rm{o}$'' and ``$\rm{a}$'' indicate objects and actions, respectively.
For $1 \leq i \leq {N}^{\rm{k}}$, the $i$-th element of $\bm{g}^{\rm{k}}$, i.e., $\bm{g}^{\rm{k}}[i] \in\{0,1\}$, indicates whether the video contains the $i$-th object (or action) or not, and our goal is to calculate the value of $\bm{g}^{\rm{k}}[i]$.
Given a video $\bm{v}$, the process of generating semantic concepts is formulated as
\begin{eqnarray}
\label{eq:scd}
     (\bm{g}^{\rm{o}},\bm{g}^{\rm{a}}) = \phi_{1}(\bm{v};\mathcal{M}_{1}),
\end{eqnarray}
where $\mathcal{M}_{1}$ represent the parameters and $\phi_{1}(\cdot)$ represent the semantic concept generator. 

To fully discover the objects and actions within a video, an object attention model and an action attention model are proposed to implement $\phi_{1}(\cdot)$. 
Let $f^{\rm{k}}(\bm{v}) \in \mathbb{R}^{q \times d}$ be the feature map of the video $\bm{v}$, the attention models $\text{ATT}^{\rm{k}}(\cdot)$ are formulated as
\begin{eqnarray}
\label{eq:attention}
     \text{ATT}^{\rm{k}}(\bm{v},\bm{e}_{i}^{\rm{k}})=(\bm{\alpha}_{i}^{\rm{k}})^{ \top} f^{\rm{k}}(\bm{v}),
\end{eqnarray}
where $\bm{e}_{i}^{\rm{k}} \in \mathbb{R}^{h \times 1}$ denotes the word embedding feature vector of the $i$-th object or action in  external sentence corpora.
$\bm{\alpha}_{i}^{\rm{k}} \in \mathbb{R}^{q \times 1}$ represents the corresponding attention coefficients, calculated as
\begin{eqnarray}
\label{eq:att_coef}
     \bm{\alpha}_{i}^{\rm{k}} = \text{softmax}\big(f^{\rm{k}}(\bm{v}) \bm{T}^{\rm{k}} \bm{e}_{i}^{\rm{k}}\big),
\end{eqnarray}
where $\bm{T}^{\rm{k}} \in \mathbb{R}^{d \times h} $ represents a transformation matrix aiming to transform the embedding feature into the visual space. 
After learning the attention coefficients, a softmax classifier is trained with the attended visual features as input, and outputs a probability distribution $\bm{p}_i^{\rm{k}}$.
For $1 \leq c \leq {N}^{\rm{k}}$, let $\bm{p}^{\rm{k}}_i[c]$ be the $c$-th element of $\bm{p}_i^{\rm{k}}$, then $\bm{g}^{\rm{k}}[i]$ is set to $1$ when $\bm{p}^{\rm{k}}_i[i]$ is larger than a threshold $\theta$, defined by
\begin{eqnarray}
\label{eq:if}
     \bm{g}^{\rm{k}}[i] = \mathbf{1}(\bm{p}_i^{\rm{k}}[i] \geq \theta),
\end{eqnarray}
where $\mathbf{1}(\cdot)$ represents an indicator function.
Only the objects and actions with $\bm{g}^{\rm{k}}[i]=1$ are used as candidate nodes to span the dependency tree.
As described above, $\mathcal{M}_{1}$ consists of the parameters of the object and action attention models and the softmax classifier.

\subsubsection{Spanning Dependency Tree}

The dependency tree describes the dependency relationships among the generated semantic concepts.
Let $\mathcal{T}$ be the generated dependency trees, the procedure of constructing dependency trees is represented as
\begin{eqnarray}
\label{eq:dtg}
     \mathcal{T} = \phi_{2}(\bm{g}^{\rm{o}},\bm{g}^{\rm{a}}),
\end{eqnarray}
where $\phi_{2}(\cdot)$ consists of two stages: root node generation and link prediction.

\paragraph{Root node generation}

The root of a dependency tree is defined as a subject-predicate phrase that comes from the semantic concept annotation.
The composition of a subject-predicate phrase is usually a noun phrase, a verb phrase, or a noun-verb phrase.
Considering that most subject-predicate phrases (root nodes) are noun-verb phrases and the generated semantic concept is either a noun phrase or a verb phrase,  we need to generate root nodes in the form of non-verb phrase.

\begin{figure}[t]
\centering
\includegraphics[width=0.98\columnwidth]{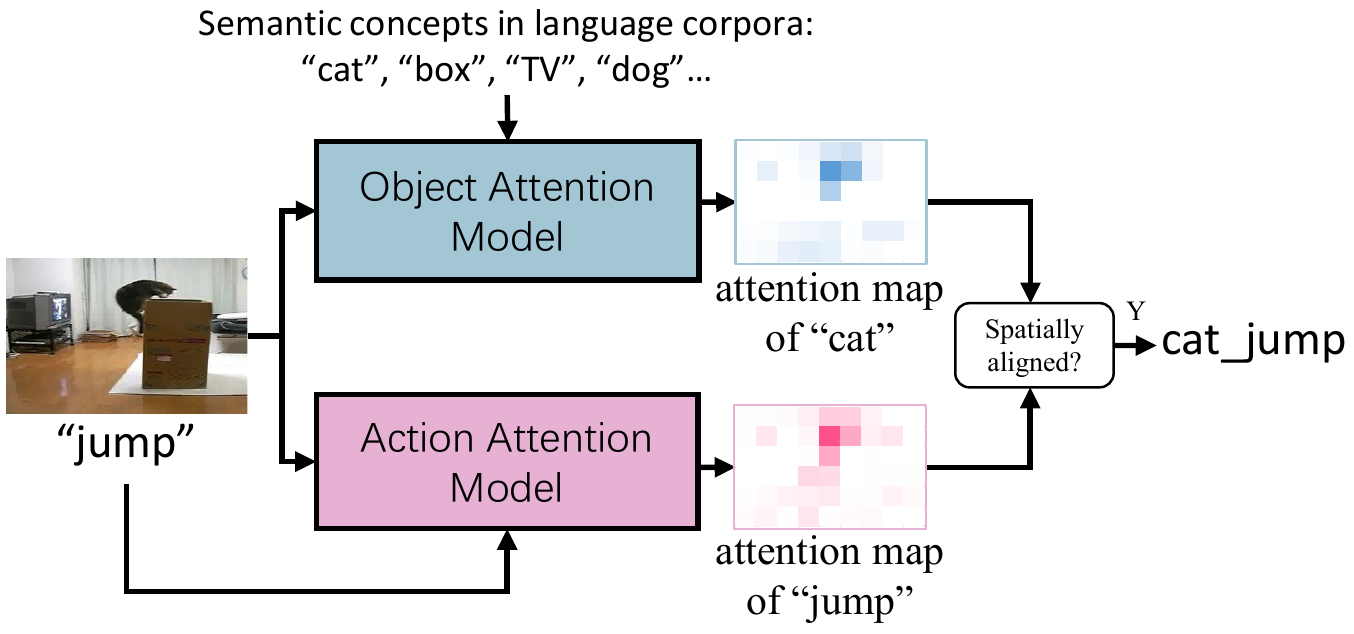}
\caption{An example of root node generation. Given a verb phase of action, i.e., ``jump'', all the candidate noun phrases in external large-scale sentence corpora are scanned. Only the noun phrase whose corresponding object is spatially aligned with the action is selected to form the root node, i.e., ``cat jump''.}
\label{rootph}
\end{figure}

Given a noun phrase or a verb phrase as a subject or a predicate, we need to infer the other corresponding phrase that might be contained in the subject-predicate phrase.
Specifically, if a verb phrase is given, all the related noun phrases among the generated semantic concepts are scanned to measure the compatibility between the noun phrase and the verb phrase according to the visual contents, and only the noun phrases with high compatibility are reserved.
If a noun phrase is given, the similar procedure is conducted on the related verb phrases to form the complete root.
Therefore, how to measure the compatibility between a noun phrase and a verb phrase is the core problem for generating the root node, where the noun corresponds to an object and the verb corresponds to an action in a video.
Since an action is often taken by an actor (referred to as an object) and has high region overlap with the actor, the attention coefficients of the actor and the action should be as similar as possible, which is defined as \emph{spatial consistency} in the object and action attention models of generating semantic concept.
The spatial consistency is determined by
\begin{eqnarray}
\label{eq:root_cond}
     &d(\bm{\alpha}_{i_{1}}^{\rm{o}},\bm{\alpha}_{i_{2}}^{\rm{a}}) \leq \delta,
\end{eqnarray}
where $\bm{\alpha}_{i_{1}}^{\rm{o}}$ represents the attention coefficient of the $i_1$-th object, $\bm{\alpha}_{i_{2}}^{\rm{a}}$ represents the attention coefficient of the $i_2$-th action, $d(\cdot)$ represents the distance between two attention coefficients, and $\delta$ is a threshold.
Figure~\ref{rootph} shows that the action of ``jump” and the object of ``cat” are spatially consistent to form the root node.

Gathering the semantic concepts as candidate nodes and root nodes, we construct the dependency trees by predicting links between the candidate nodes according to the external knowledge extracted from large-scale sentence corpora.

\paragraph{Link prediction}
Knowledge graphs (KGs) are constructed from external sentence corpora for link prediction, by using an open source toolkit, i.e., Standford CoreNLP \cite{DBLP:conf/acl/ManningSBFBM14}.
The knowledge graphs are collections of triplets, each of which represents a dependency relationship between a head entity and a tail entity.
Specifically, according to the dependency parsing, the head and tail entities can be a noun phrase, a verb phrase, or a noun-verb phrase, which are represented by the nodes in dependency trees.
The relationship denotes the preposition or whether the tail entity is the object of the head entity. 
If a sentence contains clauses or compound sentences, we simply split the sentence into multiple sub-sentences and construct multiple dependency trees accordingly.

After constructing KGs, we perform link prediction to span the dependency trees by employing a RotatE model \cite{DBLP:conf/iclr/SunDNT19}.
The score function of the RotatE model is given by
\begin{eqnarray}
\label{eq:sf}
     ||\bm{h}\circ \bm{r}-\bm{t}||_1,
\end{eqnarray}
where $\bm{h}$ and $\bm{t}$ characterize the head and tail entities, respectively.
$\circ$ is the Hadmard product operation.
$\bm{r}$ represents the relationship embedding with the constraint of $|\bm{r}(k)|=1$ where $\bm{r}(k) \in \mathbb{C}$ denotes the $k$-th element of $\bm{r}$.
Considering that some head entities in KGs might be noun-verb phrases (i.e., composition of object and action concepts), we formulate the head entity embedding by
\begin{eqnarray}
\label{eq:head}
     \bm{h} = \left\{
     \begin{array}{rcl}
     &\bm{h}^{\rm{o}},         & {\text{without} \  \bm{h}^{\rm{a}}} \\
     &\bm{h}^{\rm{a}},         & {\text{without} \  \bm{h}^{\rm{o}}} \\
     &\bm{k}_{h}\circ \bm{h}^{\rm{o}}+(1-\bm{k}_{h})\circ \bm{h}^{\rm{a}},           & {\text{otherwise}}
     \end{array} \right.
\end{eqnarray}
where $\bm{h}^{\rm{o}}$ and $\bm{h}^{\rm{a}}$ represent the head entity embeddings of object and action, respectively.
$\bm{k}_{h}$ is a gate for weighting $\bm{h}^{\rm{o}}$ and $\bm{h}^{\rm{a}}$, calculated as
\begin{eqnarray}
\label{eq:gh}
     &\bm{k}_{h} = \sigma(\bm{W}^{\top}[\bm{h}^{\rm{o}};\bm{h}^{\rm{a}};\bm{t}]+\bm{b}),
\end{eqnarray}
where $\sigma(\cdot)$ denotes the sigmoid function, $\bm{W}$ and $\bm{b}$ are parameters to be learned.
$[\bm{h}^{\rm{o}};\bm{h}^{\rm{a}};\bm{t}]$ means that the embeddings of head and tail entities are concatenated to serve as the input.
The head and tail entities are mapped into a complex domain, and the relationship $\bm{r}(k)$ has the form of $e^{i\bm{\theta}(k)}$ corresponding to a counterclockwise rotation by $\bm{\theta}(k)$ radians about the origin of the complex plane, which affects the mapping from the head entity to the tail entity in the complex vector space.

\subsubsection{Visual Reasoning and Captioning}

Given the generated dependency trees $\mathcal{T}$ as input, we use a graph convolutional network (GCN) \cite{DBLP:conf/cvpr/JohnsonGF18} coupled with a sequence-based language model \cite{DBLP:conf/cvpr/00010BT0GZ18} to reason about the relationships of the dependency trees and generate caption sentences:
\begin{eqnarray}
\label{eq:tsc}
     \mathcal{C} = \phi_{3}(\mathcal{T};\mathcal{M}_{2}, \mathcal{M}_{3}),
\end{eqnarray}
where $\mathcal{C}$ represents the output captions.
$\mathcal{M}_{2}$ and $\mathcal{M}_{3}$ represent the parameters of GCN and the sequence-based language model, respectively.
GCN contextually encodes the visual features to generate relation-aware features, which is implemented by propagating information along edges of the spanned dependency trees.
Let $\{\bm{a}_{1},\bm{a}_{2},\dots,\bm{a}_{S}\}$ represent the relation-aware features, where $S$ is the number of nodes in the dependency tree and each relation-aware feature $\bm{a}_{c}$ is generated by concatenating the feature vectors of the head entity, the relationship, and the tail entity.

To further improve the captioning performance, both the local relation-aware features and the global video features are used as the input of the language model.
The sequence-based language model is composed of a top-down attention LSTM to weigh the relation-aware features and a language LSTM to generate captions.
Specifically, the input of the top-down attention LSTM layer at time step $t$ is the concatenation of the global video feature $\bm{x}$, the embedding vector $\bm{u}_{t-1}$ of the previously generated words, and the previous hidden state $\bm{m}^2_{t-1}$ of the language LSTM layer.
Then the hidden state of the top-down attention LSTM is calculated as
\begin{eqnarray}
\begin{aligned}
\label{eq:tdalstm}
\bm{m}^1_{t} = \text{LSTM}([\bm{x};\bm{u}_{t-1};\bm{m}^2_{t-1}],\bm{m}^1_{t-1}).
\end{aligned}
\end{eqnarray}
$\bm{m}^1_{t}$ together with the relation-aware features $\{\bm{a}_{1},\bm{a}_{2},\dots,\bm{a}_{S}\}$ are employed to derive the attention weights $\bm{\beta}_t=[\beta_{t,1},\dots,\beta_{t,S}]^{\top}$, and accordingly the attended relation-aware feature is represented by $\sum_{i=1}^{G} \beta_{t,i}\bm{a}_{i}$.
The input to the language LSTM layer at time step $t$ is thus  obtained by concatenating the attended relation-aware feature with $\bm{m}^1_{t}$, and the output is the conditional probability over the words in the dictionary.
Finally, the word is generated by picking the maximum probability according to the probability and is fed into the language model together with the former input to generate the latter word.
The process is repeated until the ending token is predicted or the maximum sentence length is reached.

\subsection{Training Policy}
\subsubsection{Loss Function}
The proposed progressive visual reasoning alternates between
generating pseudo sentences and fine-tuning the video captioning model.
During the fine-tuning procedure of the captioning model, we update $\mathcal{M}_{1}$, $\mathcal{M}_{2}$ and $\mathcal{M}_{3}$ to achieve an optima in general, where $\mathcal{M}_{1}$ represents the parameters of the  semantic concept generator, $\mathcal{M}_{2}$ represents the parameters of GCN, and $\mathcal{M}_{3}$ represents the parameters of the sequence-based language model.
$\mathcal{M}_{1}$ is initialized using the ground truth semantic concepts, and $\mathcal{M}_{2}$ and $\mathcal{M}_{3}$ are randomly initialized.

During each iteration of the training procedure, we recompute the indicator variable $\bm{g}^{\rm{k}}[i]$ in Eq.(\ref{eq:if}) and rewrite it as $\widehat{\bm{g}}^{\rm{k}}[i]$ to represent the generated semantic concepts. 
$\widehat{\bm{g}}^{\rm{k}}[i]$ is calculated as
\begin{eqnarray}
\label{eq:scfinal}
     \widehat{\bm{g}}^{\rm{k}}[i] = \left\{
     \begin{array}{rcl}
     &1,        & {i \in \mathcal{I} } \\
     &\bm{g}^{\rm{k}}[i], 
     & {\text{otherwise}}
     \end{array} \right.
\end{eqnarray}
where $\mathcal{I}$ denotes the index set of generated concepts except for those generated by the attention models defined in Eq.(\ref{eq:if}).
Specifically, $\mathcal{I}$ indexes the concepts that consist of the initially given concepts $\mathcal{K}$ and the concepts appeared in the captions $\mathcal{C}$ generated in the last iteration.
Therefore, the root node is generated by integrating the generated semantic concepts and the extracted subject-predicate phrases from captions of the last iteration.
After spanning the dependency trees, a corresponding pseudo sentence is directly generated according to the language syntax to fine-tune the video captioning model.

To train the video captioning model using weak annotations, two cross-entropy losses are introduced.
One loss $L_m$ is designed to train the softmax classifier that infers the semantic concepts from the video, defined by
\begin{eqnarray}
\begin{aligned}
\label{eq:lmloss}
L_{m} = \frac{1}{N_{v}(N^{\rm{o}}+N^{\rm{a}})}\sum_{n=1}^{N_{v}}\sum_{\rm{k} \in \{\rm{o},\rm{a}\}}\sum_{i=1}^{N^{\rm{k}}}\widehat{\bm{g}}^{\rm{k}}[i]\log \bm{p}_{i}^{\rm{k}}[i],
\end{aligned}
\end{eqnarray}
where $N_{v}$ is the number of training videos. $N^{\rm{o}}$ and $N^{\rm{a}}$ represent the total numbers of objects and actions in external sentence corpora, respectively. $\widehat{\bm{g}}^{\rm{k}}[i]$ is generated via Eq.~(\ref{eq:scfinal}).
The other loss $L_c$ is introduced to train the softmax classifier in the sequence-based language model to generate the words in the captions, formulated by
\begin{eqnarray}
\begin{aligned}
\label{eq:lcloss}
L_{c}=-\frac{1}{N_{p}}\sum_{n=1}^{N_{p}}\sum_{t=1}^T\log \big( Pr(w_t|w_{1:t-1},\bm{x},\bm{s})\big),
\end{aligned}
\end{eqnarray}
where $Pr(w_t|w_{1:t-1},\bm{x},\bm{y})$ denotes the probability that the prediction is the ground-truth word $w_t$ given the previous word sequence $w_{1:t-1}$, the attended feature $\bm{s}=\text{ATT}(\bm{v},\bm{e})$ of the video $\bm{v}$ and the word embedding feature $\bm{e}$, and the global video feature $\bm{x}$.
$N_{p}$ represents the number of pairs of video and pseudo sentence, and $T$ is the total number of words in each pseudo sentence.
Taken together, all the loss functions mentioned above form the complete objective:
\begin{eqnarray}
\begin{aligned}
\label{eq:seloss}
L = \lambda L_{m} + L_{c} + \|\mathcal{M}\|_{2},
\end{aligned}
\end{eqnarray}
where $\lambda$ is a hyper-parameter and $\mathcal{M} = \mathcal{M}_{1} \cup \mathcal{M}_{2} \cup \mathcal{M}_{3}$ represent all the parameters to be learned. 

\subsubsection{Iterative Refinement Algorithm}

Algorithm \ref{alg:alg1} illustrates the iterative refinement algorithm of our progressive visual reasoning method.
The training data set $\mathcal{P}$ is represented by $\{(\bm{v},[\widehat{\bm{g}}^{\rm{o}},\widehat{\bm{g}}^{\rm{a}}],y)\}$ where $\bm{v} \in \mathcal{V} $ represents a training video and $y \in \mathcal{Y}$ represents its corresponding pseudo sentence.
$[\widehat{\bm{g}}^{\rm{o}},\widehat{\bm{g}}^{\rm{a}}]$ indicates the concatenation of $\widehat{\bm{g}}^{\rm{o}}$ and $\widehat{\bm{g}}^{\rm{a}}$ that are calculated via Eq.(\ref{eq:scfinal}).
Initially ($flag = 1$), the object and action attention models are learned using weak annotations.
Then the learned attention models infer semantic concepts to provide candidate nodes of dependency trees, and the pseudo sentences generated from the dependency trees are used to fine-tune the video captioning model.
Finally, the generation of pseudo sentences and the fine-tuning of video captioning model are alternated until reaching the convergence, i.e., no more new pseudo sentences are generated or the evaluation metric CIDEr on the validation set is not increasing.

\IncMargin{1em}
\begin{algorithm}
\SetKwInOut{INIT}{Initialization}
\small
\caption{Iterative Refinement Algorithm of Progressive Visual Reasoning for Video Captioning.}
\label{alg:alg1}
\KwIn{
Training videos $\mathcal{V}$ and annotated semantic concepts $\mathcal{K}$.
}
\KwOut{Captioning model.} 
$\bullet$ \textbf{Initializations} \\
\ \ \ $\mathcal{C} \leftarrow \emptyset, \mathcal{Y} \leftarrow \emptyset$,$\mathcal{M} \leftarrow \emptyset$,$N_{v} \leftarrow |\mathcal{V}|$; \\
\ \ \ Initialize training data $\mathcal{P} \leftarrow \emptyset$; \\
\ \ \ Initialize $\widehat{\bm{g}}^{\rm{o}}$ and $\widehat{\bm{g}}^{\rm{a}}$ according to $\mathcal{K}$; \\
\ \ \ $flag \leftarrow 1$; \\
\Repeat{Convergence}
{

    $\bullet$ \textbf{Generate Pseudo Sentence} \\
    \If{flag $\neq 1$}{
    Generate root nodes for each tree by using $\widehat{\bm{g}}^{\rm{o}}$, $\widehat{\bm{g}}^{\rm{a}}$, $\mathcal{K}$, and $\mathcal{C}$; \\
    Predict links of trees to obtain pseudo sentences $\mathcal{Y}$ for each video; \\
    }
    $\mathcal{P} \leftarrow \mathcal{P} \cup \{(\bm{v},[\widehat{\bm{g}}^{\rm{o}},\widehat{\bm{g}}^{\rm{a}}],y)|y \in \mathcal{Y} \cup \mathcal{C}\}$; \\
    $N_{p} \leftarrow |\mathcal{P}|$; \\
    $\bullet$ \textbf{Fine-tune Video Captioning Model} \\
    Update $\mathcal{M}$ with $\mathcal{P}$ by minimizing Eq.~(\ref{eq:seloss}); \\
    \If{flag $\neq 1$}{
    Calculate captions $C$ via the updated captioning model for each video; \\
    }
    Update $\hat{\bm{g}}^{\rm{o}}$ and $\hat{\bm{g}}^{\rm{a}}$ via Eq.~(\ref{eq:scfinal}); \\
    $flag \leftarrow 0$;
}
\end{algorithm}

\begin{table*}[htbp]
  \small
  \centering
  \caption{Comparison results of different sentence corpora on the MSVD and MSR-VTT datasets.}
    \resizebox{0.85\textwidth}{!}{
    \begin{tabular}{l|cccc|ccccc}
    \toprule
    \multirow{2}[4]{*}{Sentence corpora} & \multicolumn{4}{c|}{MSVD}     & \multicolumn{4}{c}{MSR-VTT} \\
\cmidrule{2-9}          & B@4    & METEOR  & ROUGE-L & CIDEr & B@4    & METEOR  & ROUGE-L & CIDEr \\
    \midrule
    MSVD & \textbf{47.2}  & 33.9  & \textbf{70.9}   & 68.9   &  33.1  & 26.2  &  59.1 & 38.7 \\
    MSR-VTT & 46.1 & 29.6 & 62.4 & 62.4  & 39.1 & \textbf{27.5} & 59.8  &  40.1 \\
    MSVD+MSR-VTT & \textbf{47.2} &   \textbf{34.3}    & \textbf{70.9} &  \textbf{72.1}   &\textbf{39.9} & \textbf{27.5} & \textbf{59.9} & \textbf{40.3} \\
    MS-COCO & 32.1 & 29.2 & 60.6 &  36.4 &  28.1 & 25.7 & 58.1 & 35.3 \\
    \bottomrule
    \end{tabular}}%
  \label{tab:sentc}%
\end{table*}%

\section{Experiments}

\subsection{Datasets and Sentence Corpora}
To quantitatively evaluate our method, we conduct experiments on two video captioning datasets: MSVD \cite{DBLP:conf/acl/ChenD11} and MSR-VTT  \cite{DBLP:conf/cvpr/XuMYR16}.
\begin{itemize}
\item The MSVD dataset comprises 1,970 videos collected from Youtube, each of which has roughly 40 captions.
Following \cite{DBLP:conf/iccv/VenugopalanRDMD15}, we split the videos into three sets: a training set of 1,200 videos, a validation set of 100 videos and a testing set of 670 videos.

\item The MSR-VTT dataset contains 10,000 video clips, where each video is annotated with 20 captions.
Following \cite{DBLP:conf/cvpr/XuMYR16}, we take 6,513 videos for training, 497 videos for validation, and 2,990 videos for testing.

\end{itemize}


We randomly extract a noun phrase or a verb phrase that describes an object or an action from the ground-truth sentences of each training video as the weak annotation.
In order to avoid the confusion when classifying different words with overlapping semantics, such as words of ``cat'', ``dog'', and ``animal'', we use WordNet\footnote{\url{https://wordnet.princeton.edu}}to remove words that are hypernyms of other words in the dictionary, such as ``animal''.
By cutting down the number of words in the dictionary, the computational complexity can be relatively reduced.
In total, videos in MSVD are labeled with 396 object classes and 168 action classes, and videos in MSR-VTT are labeled with 846 object classes and 382 action classes.

We directly use the ground-truth sentences of MSVD and MSR-VTT to construct the different sentence corpora  to avoid the impact on the evaluation metrics caused by different linguistic characteristics \cite{DBLP:journals/corr/abs-1811-10787} of the ground-truth test sentences in the two datasets and the sentences in the external corpus.
In addition, we conduct experiments in an unpaired manner using the sentences in the MSCOCO dataset \cite{lin2014microsoft} as the sentence corpus to simulate the realistic scenario.


\subsection{Implementation Details}

InceptionResNet-v2  \cite{DBLP:conf/aaai/SzegedyIVA17} is used to extract visual features as video representation, and the embedded vectors of Glove \cite{pennington2014glove} are utilized to represent words.
Specifically, the feature maps extracted from the Reduction-C layer of InceptionResNet-v2 are used as input video features to the object and action attention models. 
The 300-dimensional embedded vectors in Glove are taken as input word features to the object and action attention models. 
To further improve the captioning performance, we combine
the features extracted from the average pooling layer of InceptionResNet-v2 and  the pool5 layer of C3D \cite{tran2015learning} as the global video feature $\bm{x}$ of the sequence-based language model.

In the attention models, the dimension $d$ of the output attended feature is set to 512.
In GCN, the parameter settings are the same as \cite{DBLP:conf/cvpr/JohnsonGF18}, and the parameter settings of the sequence-based language model are the same as  \cite{DBLP:conf/cvpr/00010BT0GZ18}.
The size of the beam search in the sequence-based language model is set to $5$.
The hyper-parameter $\lambda$ in Eq.~(\ref{eq:seloss}) is set to 0.1.
The thresholds $\theta$ in Eq.~(\ref{eq:if}) and $\delta$ in Eq.~(\ref{eq:root_cond}) are set to 0.99 and 0.1, respectively.
The RMSprop \cite{DBLP:journals/corr/Graves13} is employed to optimize our model and the learning rate is set to $1{\rm e}^{-4}$.

We use the metrics of BLEU-4 (B@4) \cite{DBLP:conf/acl/PapineniRWZ02}, METEOR \cite{DBLP:conf/wmt/DenkowskiL14}, ROUGE-L \cite{lin2004rouge} and CIDEr \cite{DBLP:conf/cvpr/VedantamZP15} calculated by the MSCOCO toolkit \cite{DBLP:journals/corr/ChenFLVGDZ15} for evaluation.
For all the metrics, higher values indicate better performances.

\subsection{Effect of Knowledge from Different Sentence Corpora}
To evaluate the effect of knowledge extracted from different sentence corpora, we compare the results of our method on the MSVD and MSR-VTT datasets using four different sentence corpora.
The sentence corpora are collected from the training sentences in the video captioning datasets (MSVD and MSR-VTT) and all the sentences in the image captioning dataset (MSCOCO). The four different sentence corpora are: MSVD, MSR-VTT, MSVD+MSR-VTT and MSCOCO.

Table~\ref{tab:sentc} shows the comparison results of four different sentence corpora on both MSVD and MSR-VTT datasets. From the results,  we can have the following observations:
(1) The larger sentence corpora from the video datasets (MSVD+MSR-VTT) achieves the best results for most metrics, demonstrating that a large corpus can provide more abundant dependency relationships between semantic concepts to generate more accurate pseudo sentences for training; (2) The corpus from the image dataset (MSCOCO) generally works worse than other corpus from the video datasets, probably due to that the sentences for describing still images are not sufficient to express the motion information in videos.

In the following experiments, we collect  the training sentences of the MSVD and MSR-VTT datasets as the external sentence corpora (MSVD+MSR-VTT) due to its superior performance.
\subsection{Effect of Different Training Data }
Since there exist many public available video datasets with weak annotations (e.g., action recognition datasets with action class labels), we use different combinations of video captioning datasets (MSVD and MSR-VTT with weak annotations) and action recognition dataset (HMDB51 \cite{kuehne2011hmdb}) as training data.
The HMDB51 dataset contains 6,766 video clips annotated with 51 action labels, and the weak annotations for these videos are the action class labels.



\begin{table}[htbp]
  \small
  \centering
  \caption{Comparison results of different training videos on the MSVD dataset where the external sentence corpora comes from the training sentences of the MSVD and MSR-VTT datasets.}
    \resizebox{0.47\textwidth}{!}{
    \begin{tabular}{l|cccc}
    \toprule
    Training videos & B@4 & METEOR  & ROUGE-L & CIDEr \\
    \midrule
    MSVD  &47.2 &   34.3    & {70.9} &  {72.1}  \\
    MSVD+MSR-VTT & 47.3 & 34.1 & 71.1 & 77.2 \\
    MSVD+MSR-VTT+HMDB51 &\textbf{47.9}  &  \textbf{34.7} & \textbf{71.3} & \textbf{80.1} \\
    \bottomrule
    \end{tabular}}%
  \label{tab:datasets_msvd}%
\end{table}%

\begin{table}[htbp]
  \small
  \centering
  \caption{Comparison results of different training videos on the MSR-VTT dataset where the external sentence corpora comes from the training sentences of the MSVD and MSR-VTT datasets.}
    \resizebox{0.47\textwidth}{!}{
    \begin{tabular}{l|cccc}
    \toprule
    Training videos & B@4 & METEOR  & ROUGE-L & CIDEr \\
    \midrule
    MSR-VTT &39.9 & 27.5 & 59.9 & 40.3 \\ 
    MSVD+MSR-VTT & 40.1 & 27.7 & \textbf{60.1} & 43.3 \\
    MSVD+MSR-VTT+HMDB51 & \textbf{40.4} & \textbf{27.9} & \textbf{60.1}  & \textbf{45.3} \\
    \bottomrule
    \end{tabular}}%
  \label{tab:datasets_msrvtt}%
\end{table}%

For evaluation on the MSVD test set, we train our model using the training videos of the following datasets: (1) MSVD; (2) combination of  MSVD and MSR-VTT (MSVD+MSR-VTT); (3)  combination of  MSVD, MSR-VTT and HMDB51 (MSVD+MSR-VTT+ HMDB51).
Table~\ref{tab:datasets_msvd} shows the comparison results of our method on the MSVD test set using these different training sets.
Similarly, on the MSR-VTT test set, we use the training videos of the following datasets: (1) MSR-VTT; (2) MSVD+MSR-VTT; (3) MSVD+MSR-VTT+HMDB51. The comparison results are shown in Table~\ref{tab:datasets_msrvtt}.
Note that for the results shown in Table ~\ref{tab:datasets_msvd} and Table~\ref{tab:datasets_msrvtt}, the external sentence corpora comes from the training sentences of the MSVD and MSR-VTT datasets.
From the results, it is interesting to observe that the performances on both the MSVD and MSR-VTT  datasets  improve  with the increasing number of training videos, demonstrating the feasibility and the effectiveness of training videos with weak annotations on learning the video captioning model.


\subsection{Analysis on Iterative Refinement}
\begin{figure}[t]
\centering
\includegraphics[width=1\columnwidth]{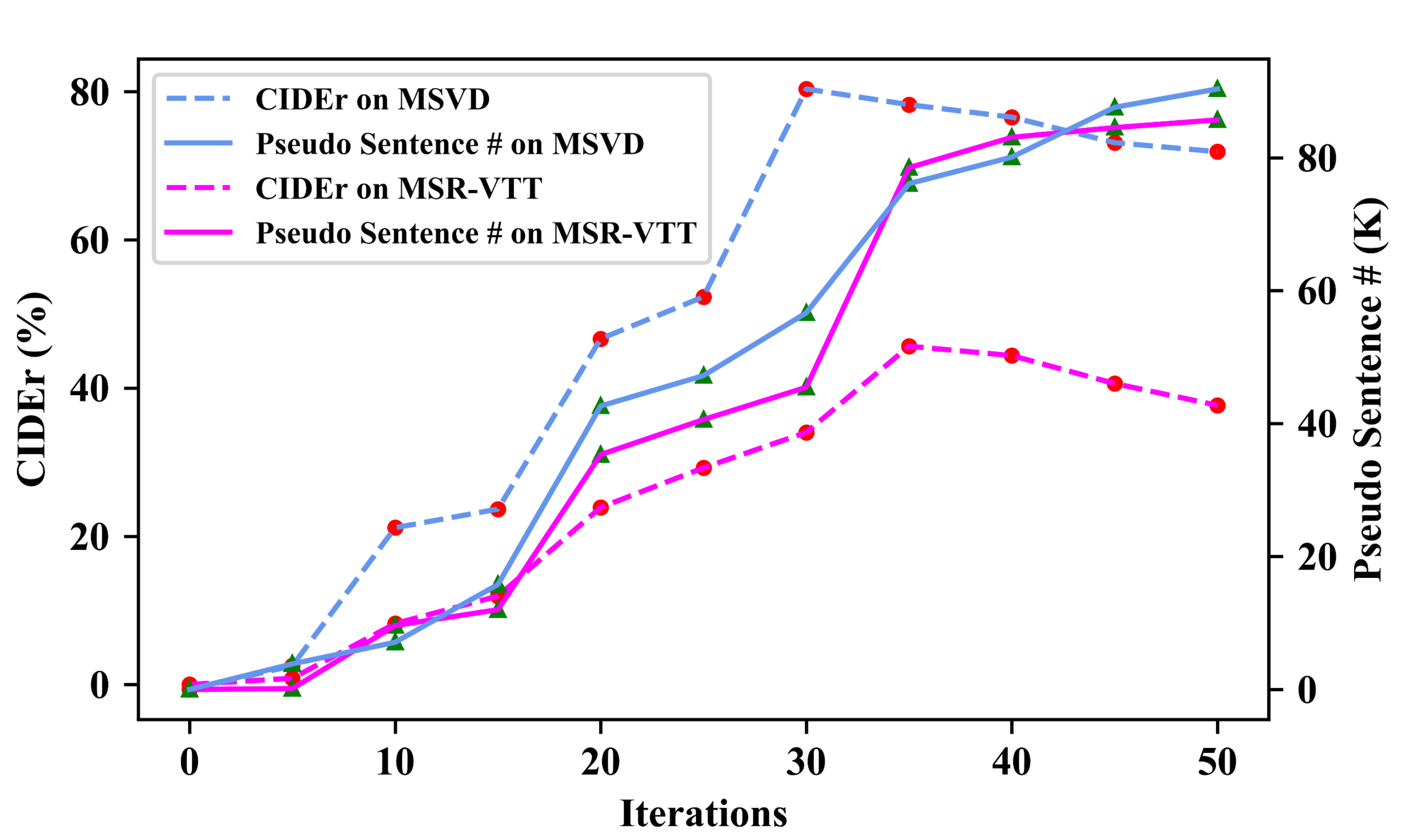} 
\caption{The trends of the CIDEr score and the number of  generated pseudo sentences with increasing iterations on the MSVD and MSR-VTT datasets.}
\label{exp}
\end{figure}

Figure~\ref{exp} shows the CIDEr scores and the numbers of generated pseudo sentences at each iteration on the MSVD and MSR-VTT datasets. 
We can see that the CIDEr score gradually increases, which demonstrates the effectiveness of the iterative refinement strategy on video captioning using weak annotation.
The trend of the generated pseudo sentence number is consistent with the trend of the CIDEr score, indicating that it is important to generate more diverse pseudo sentences for training.
Moreover, when the number of generated pseudo sentences grows slowly, the CIDEr score decreases, probably due to the over-fitting problem in training the video captioning model using limited pseudo sentences. 




\begin{table*}[htbp]
  \small
  \centering
  \caption{Comparison results  of our method and different state-of-the-art methods on the MSVD and MSR-VTT datasets.}
  \resizebox{0.86\textwidth}{!}{
    \begin{tabular}{l|c|cccc|cccc}
    \toprule
    \multirow{2}{*}{Methods}&\multirow{2}{*}{Supervision} & \multicolumn{4}{c|}{MSVD}     & \multicolumn{4}{c}{MSR-VTT} \\
\cmidrule{3-10}    &      & B@4    & METEOR  & ROUGE-L & CIDEr & B@4    & METEOR  & ROUGE-L & CIDEr \\
    \midrule
    Venugopalan et al. \cite{DBLP:conf/naacl/VenugopalanXDRM15}&\multirow{9}{*}{Strong} & 33.3  & 29.1  & -     & -     & -     & -     & -     & -\\
    Yao et al. \cite{DBLP:conf/iccv/YaoTCBPLC15} && 41.9  & 29.6  & -     & 51.7  & 37.1  & 28.4  & -     & - \\
    Yu et al. \cite{DBLP:conf/cvpr/YuWHYX16} && 49.9  & 32.6  & -     & 65.8  & -     & -     & -     & -\\
    Gao et al. \cite{DBLP:journals/tmm/GaoGZXS17} && 50.8  & 33.3  & 61.1  & 74.8  & 38.0  & 26.1  & -     & 43.2  \\
    Chen et al. \cite{Chen_2018_ECCV} && 52.3  & 33.3  & 69.9  & 76.5  & 41.3  & 27.7  & 59.8  & 44.1  \\
    Liu et al. \cite{DBLP:conf/mm/LiuRY18} && 54.2  & 34.8  & 71.7  & 88.2  & 40.9  & 27.5  & 60.2  & 47.5  \\
    Aafaq et al. \cite{DBLP:journals/corr/abs-1902-10322} && 47.9  & 35.0  & 71.5  & 78.1  & 38.3  & 28.4  & 60.7  & 48.1  \\
    Zhang et al. \cite{DBLP:journals/corr/abs-1906-04375} && 56.9  & 36.2  & -     & 90.6  & 41.4  & 28.2  & -     & 46.9  \\
    Zhang et al. \cite{zhangcvpr2020caption} && 54.3  & 36.4  & 73.9     & 95.2  & 43.6  & 28.8  & 62.1     & 50.9  \\
    \midrule
    Feng et al. \cite{DBLP:journals/corr/abs-1811-10787}  &No& 15.3  &  20.5 & 57.7 & 21.6   & 22.1 & 22.6 & 51.4 & 13.5 \\
    \midrule
    Ours  &Weak& 47.9  &  34.7 & 71.3 & 80.1   & 40.4 & 27.9 & 60.1  & 45.3 \\
    \bottomrule
    \end{tabular}}%
  \label{tab:strong}%
\end{table*}%

\subsection{Comparison with strongly-supervised and unsupervised methods}
\subsubsection{Comparison with strongly-supervised methods.}
Table~\ref{tab:strong} reports results of our method and the strongly-supervised methods~\cite{DBLP:conf/naacl/VenugopalanXDRM15,DBLP:conf/iccv/YaoTCBPLC15,DBLP:conf/cvpr/YuWHYX16,DBLP:journals/tmm/GaoGZXS17,Chen_2018_ECCV,DBLP:conf/mm/LiuRY18,DBLP:journals/corr/abs-1902-10322,DBLP:journals/corr/abs-1906-04375,zhangcvpr2020caption} (i.e., methods using strong annotation) on the MSVD and MSR-VTT datasets. Note that the recent state-of-the-art strongly-supervised methods~ \cite{DBLP:journals/corr/abs-1906-04375,zhangcvpr2020caption,DBLP:journals/corr/abs-1902-10322} additionally use object detectors to further improve the captioning performance. 
From Table~\ref{tab:strong}, it is interesting to notice that although our method uses merely weak annotations for training, it still achieves promising results and is competitive to the strongly-supervised methods, which validates the effectiveness of the progressive visual reasoning.

\subsubsection{Comparison with unsupervised methods.}
We also compare our method with the unsupervised method \cite{DBLP:journals/corr/abs-1811-10787} in Table~\ref{tab:strong}, since \cite{DBLP:journals/corr/abs-1811-10787} also uses semantic concepts to generate captions.
Our method performs much better probably due to the following reasons:
(1) \cite{DBLP:journals/corr/abs-1811-10787} detects semantic concepts via a detector pre-trained on a large-scale image dataset, while our method infers semantic concepts from weak annotations and would not introduce bias in training the captioning model.
(2) Rather than directly reconstructing images and sentences in \cite{DBLP:journals/corr/abs-1811-10787}, we explicitly exploit the relationships between semantic concepts to  gradually generate refined pseudo sentences to train the captioning model with good interpretability and   generalization ability. 

\begin{figure*}[t]
\centering
\includegraphics[width=0.86\textwidth]{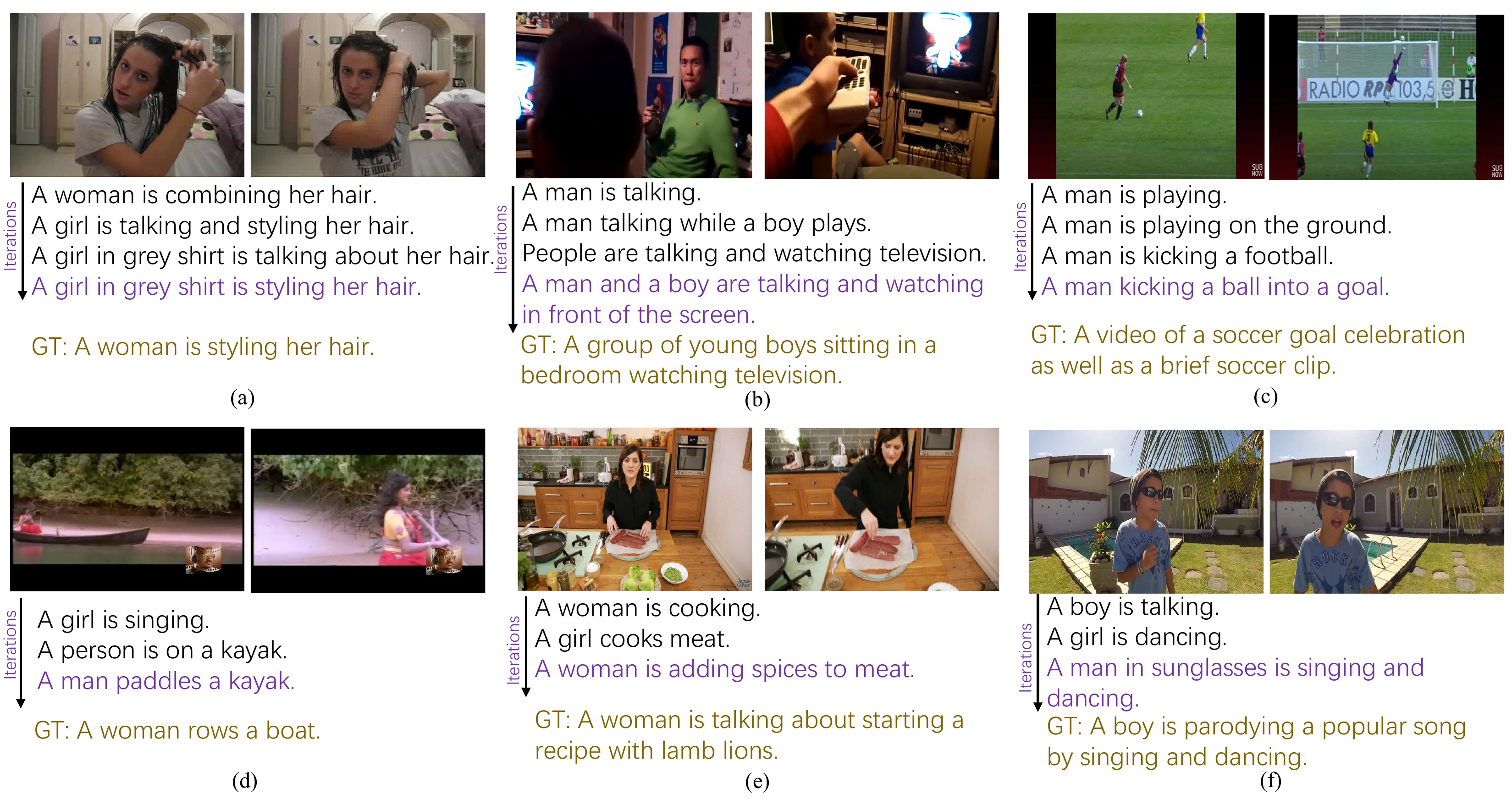} 
\caption{Qualitative results of video captioning on the MSVD and MSR-VTT datasets. The generated captions through iterations are shown under the corresponding video frames. The ground-truth (GT) captions are presented at the bottom line.}
\label{qua}
\end{figure*}

\subsection{Qualitative Analysis}

Figure~\ref{qua} shows some qualitative results of the test data in the MSVD and MSR-VTT datasets.
For each video, two frames are randomly sampled to represent the video.
We sample the output captions inferred by the captioning model from the beginning of the training to the end with a fixed step during the iteration process.
It is interesting to observe that the generated captions by our method are gradually refined through the progressive visual reasoning.
Taking Figure~\ref{qua} (b) as an example, in the first few iterations of our iterative refinement algorithm, our captioning model only generates ``a man is talking'' containing only two concepts, which is too general to precisely describe the video.
Benefiting from the progressive reasoning mechanism, our method is able to discover more concepts to enable more precise captions to be generated.

\section{Conclusion}

We have presented a progressive visual reasoning method for video captioning using weak annotation.
Inspired by the curriculum learning, our method can exploit external knowledge to reason about relationships between semantic concepts along dependency trees and can progressively expand a single semantic concept to pseudo sentences for each training video.
By using the generated pseudo sentences, the video captioning model can be trained well to achieve promising results. 
The experimental results demonstrate that our method using weak annotation is very competitive to the state-of-the-art methods using strong annotation.

\bibliographystyle{IEEEtran}
\bibliography{acmart}

\end{document}